\documentclass[lettersize,journal]{IEEEtran}
\usepackage{amsmath,amsfonts}
\usepackage{algorithmic}
\usepackage{algorithm}
\usepackage{array}
\usepackage[caption=false,font=normalsize,labelfont=sf,textfont=sf]{subfig}
\usepackage{textcomp}
\usepackage{stfloats}
\usepackage{url}
\usepackage{verbatim}
\usepackage{graphicx} 
\usepackage{cite}
\usepackage[numbers,sort&compress]{natbib} 
\usepackage{booktabs}
\usepackage{caption}
\usepackage{subcaption} 
\usepackage[T1]{fontenc}

\begin{document}

\title{An Interpretable Deep Learning Framework for Material Perception and Classification from Multisensory Tactile Data}

\author{Li Zou, Dave Hogendoorn, Yasemin Vardar
\thanks{Authors are with the Department of Cognitive Robotics, Faculty of Mechanical Engineering, Delft University of Technology, Delft, The Netherlands.}
\thanks{This work is supported by ERC grant (SuperTouch, 101220242).}
\thanks{Corresponding author: Yasemin Vardar}}

\maketitle

\begin{abstract}
Human tactile perception relies on complex multisensory cues, yet the relationship between tactile signals and perceptual representations remains poorly understood, limiting touch integration in digital environments and human-like robotic perception. To address this gap, we developed a computational framework comprising three interconnected deep learning models that map multisensory touch data to material perception, without relying on hand-crafted features. The models represent progressively different routes from tactile signals to material class: from low-level interaction signals to perceptual attribute distributions (Model 1), from predicted attribute distributions to material classification (Model 2), and directly from tactile signals to material categories, bypassing intermediate representations (Model 3). By combining deep learning with Integrated Gradients, the framework achieved high accuracy while offering interpretability, revealing which sensory modalities most strongly drive its decisions. Our results show that deep learning can approach near-perfect material classification when unconstrained by intermediate perceptual stages, but matching human-like performance is harder once those stages are modeled explicitly. Notably, thermal cues emerged as particularly informative across all models, providing robust signals for material differentiation. The results offer a computational account of how tactile signals lead to material perception and show how interpretable deep learning can both approach human-level performance and reveal cues that robotic and haptic systems need to incorporate. 

\end{abstract}


\section{Introduction}
\IEEEPARstart{T}{ouch} is central to how humans perceive and engage with the physical world—whether in effortless actions like gripping a coffee cup or a spoon, or in more deliberate haptic exploration when evaluating the texture of a surface. Each instance of contact generates an intricate pattern of tactile signals in the skin, which the nervous system decodes to infer key perceptual qualities, such as roughness, slipperiness, and hardness, enabling robust and rapid material recognition. These signals arise from distinct exploratory actions---including pressing, sliding, and static contact---each providing complementary information about different facets of surface properties, from mechanical features like friction, compliance, and roughness, to thermal characteristics~\cite{klatzky2025_action, lederman1987_hand, callier2015_kinematics}. Together, they form a rich sensory tapestry that underpins our nuanced interaction with objects and environments.

Despite humans' exceptional ability to integrate tactile signals, the processes by which tactile information is transformed into meaningful perceptual attributes—and ultimately into material recognition—remain poorly understood. Key open questions include which sensory cues are most informative, how their combination shapes perception, and how they give rise to distinct tactile sensations. This gap in understanding limits the development of digital interfaces that can faithfully reproduce touch and hinders the design of robotic systems capable of human-like tactile perception.

Machine learning has recently become a powerful tool for analyzing complex datasets and has found widespread application in material classification. One common strategy uses supervised learning with hand-engineered features extracted from tactile or multimodal signals~\cite{Strese2020_exploratory, Devillard2025_database, Zackory2017_semisupervised, Shuvo2025_classification, fishel2012_bayesien, zou2026}. These methods remain popular due to their relative simplicity and interpretability, and have yielded classification accuracies between 74\% and 95\%, especially when combining multisensory tactile signals with additional modalities like vision or audition. Despite its utility, this supervised approach is constrained by its reliance on predetermined, hand-crafted features: deriving these representations demands considerable domain expertise and entails a tedious, subjective design process, which ultimately limits the model's adaptability.

When the mapping from measurements to outcomes is complex, learning it directly from data is often more effective than manually encoding rules for every possible scenario. Deep learning facilitates this by automatically learning useful features from raw, high-dimensional signals, and has proven highly effective across many perceptual domains~\cite{LeCun2015}, achieving classification accuracies as high as 98.8\% on tactile recognition tasks~\cite{saga2020_classification, Zhang2024_visuotactile, Zheng2016_deeplearning}. 

Despite these developments, the interdependencies among multimodal tactile signals, intermediate perceptual descriptors, and high-level material categories remain theoretically and empirically underdetermined. Studies employing multisensory tactile data have commonly optimized for discriminative power~\cite{Strese2020_exploratory, Strese2017_multimodal}, without modeling the perceptual attributes that mediate material judgments. Perception-focused work has typically operated within reduced parameter spaces—often limited to frictional or vibratory data~\cite{richardson2020_learningtopredict, richardson2022_learn2feel, lim2025_machine}. Other work has examined how perceptual attributes (e.g., roughness, slipperiness, hardness) relate to material judgments without modeling how these attributes emerge from underlying tactile signals~\cite{Baum2013_visual}. More recently, \cite{zou2026} attempted to capture these interdependencies using hand-crafted features, but this approach limited the model's interpretability.

To bridge this gap, we leverage a computational framework proposed by \cite{zou2026} for modeling human material perception and recognition, extending it by replacing hand-crafted features and classical machine learning methods with deep learning throughout the entire pipeline. The framework operates on multisensory touch signals (force, acceleration, indentation, temperature, heat flux) collected across various exploratory actions (sliding, pressing, and static contact), and can be decomposed into two interrelated but distinct facets: algorithmic and structural.

At the algorithmic level, the framework integrates deep learning with post-hoc interpretability analysis to identify materials and to uncover the decision process of the deep learning algorithms—specifically, which signal modalities and which temporal segments within a signal contribute most to the classification.

\begin{figure}[t!]
    \centering
    \includegraphics[width=0.8\linewidth, trim={170 160 280 100}, clip]{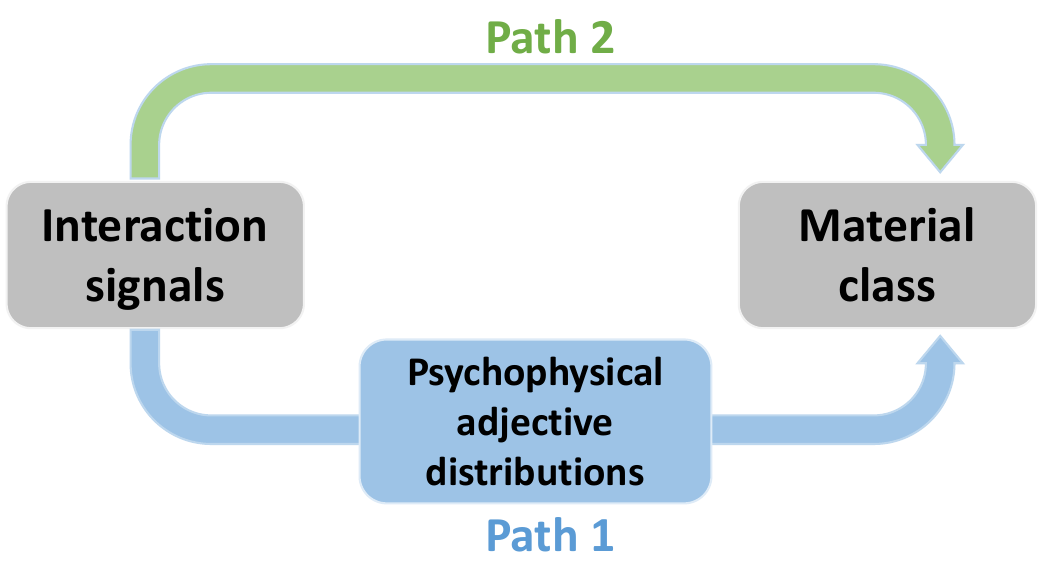}
    \caption{The computational framework for modeling human material perception and recognition via interaction signals.}
    \label{fig:model_overview}
\end{figure}

At the structural level, the framework comprises two pathways built from three interconnected models (Fig.~\ref{fig:model_overview}). Path~1 proceeds from finger–surface interaction signals to psychophysical sensory attributes (Model 1) and then to material classes (Model 2), probing the AI's capacity for human-aligned haptic perception. Path~2 instead maps finger–surface interaction signals directly to material classes, bypassing any intermediate stage (Model 3), without such alignment. This bi-pathway design enables a comparison of how AI models perform haptic material recognition with and without alignment to human perception.

\section{Methods}

\subsection{Dataset}

For tactile signals, we utilized SENS3, an open-access multisensory dataset collected from fifty different surfaces~\cite{balasubramanian2025_sens3}. Tactile data were recorded using an apparatus equipped with a 3D accelerometer, a 6D force-torque sensor, a thermistor, a heat-flux sensor, and an infrared position sensor. Signals were collected while three participants explored each surface with their fingertips using various exploratory actions, including static contact, pressing, and sliding. 

Pressing produced force and indentation depth signals as the index finger applied normal pressure to each surface. Static contact captured thermal data, including heat flux and skin temperature. Sliding yielded lateral force and acceleration signals characterizing surface friction and roughness. For interpretability analysis, static contact was divided into peak, half-equilibration, and steady phases; pressing into loading, plateau, and lift-off; and sliding into 12 force–speed bins covering 0.2–0.8~N and 33–165~mm/s, divided into three and four ranges, respectively. Further details regarding the recording procedure can be found in \cite{balasubramanian2025_sens3}.

SENS3 comprises 50 materials distributed across 10 classes. To ensure statistical robustness in the classification task, we excluded classes with two or fewer material instances, resulting in a dataset of 45 materials across seven classes: fabric, foam, metal, paper, sandpaper, vinyl, and wood.

SENS3 also provides psychophysical ratings from a separate experiment, where 20 participants rated each material on adjective pairs like hot--cold, hard--soft, rough--smooth, and sticky--slippery using a 1--15 discrete scale~\cite{balasubramanian2025_sens3, zou2026}. These adjective pairs represent the primary psychophysical dimensions of tactile surface perception found in the literature~\cite{okamoto2013_dimensions, Baum2013_visual, balasubramanian2025_sens3}.

To mitigate rater bias, we centered each participant's ratings on the global scale mean. Let $r_{p,m,a}$ be the rating from participant $p$ for material $m$ and adjective pair $a$; let $\bar{r}_{p,a}=\frac{1}{45}\sum_{m=1}^{45}r_{p,m,a}$ denote the participant's mean rating across all materials for adjective pair $a$; and let $\bar{r}_{\mathrm{global}} = 8$ the scale midpoint. Each rating was adjusted as
\begin{equation}
r^{\mathrm{adj}}_{p,m,a} = \mathrm{clip}\left(r_{p,m,a} - \bar{r}_{p,a} + \bar{r}_{\mathrm{global}},\, 1,\, 15\right),
\end{equation}
which preserved relative ordering while removing systematic rater biases. For each material $m$ and adjective pair $a$, we aggregated adjusted ratings over 20 participants into counts $c_k(m,a)$, denoting the number of ratings falling in the interval $(k-1, k]$. A small smoothing constant $0.1$ was added to avoid empty bins:
\begin{equation}
\tilde{c}_k(m,a) = c_k(m,a) + 0.1,
\end{equation}
and normalize to obtain a probability distribution:
\begin{equation}
p_k(m,a) = \frac{\tilde{c}_k(m,a)}{\sum_{j=1}^{15} \tilde{c}_j(m,a)}.
\end{equation}
This operation yielded one empirical distribution $p(m,a) \in \Delta^{15}$ per material and adjective pair, which served as the training target for the adjective-prediction model (Model~1).

\subsection{1D-Convolutional Neural Networks (1D-CNNs)}

A 1-Dimensional Convolutional Neural Network (1D-CNN) processes sequential data (e.g., time series) by sliding a learnable kernel along one dimension~\cite{Ige2024Survey1D, Cacciari2024HandsOn1DCNN}. For a multichannel input $\mathbf{x} \in \mathbb{R}^{T \times C}$ and kernel $\mathbf{W}^{(k)} \in \mathbb{R}^{K \times C}$ with bias $b^{(k)}$, the $k$-th feature map at time $t$ is given by
\[
y^{(k)}_t = \sigma\!\left(\sum_{i=0}^{K-1} \sum_{c=1}^{C} W^{(k)}_{i,c} \, x_{t+i-\Delta,\,c} + b^{(k)}\right),
\]
where $T$ is the sequence length, $K$ is kernel width, $\Delta$ sets the padding offset, $\sigma$ is a non-linearity (e.g., ReLU), $C$ is the number of channels, and $k$ indexes the kernels used.

Compared to recurrent or hand-crafted features, 1D-CNNs offer weight sharing, stable parallel training, and precise control over the receptive field via kernel size, stride, and dilation. These properties make them well suited to thermo-tactile signals, where brief high-amplitude events (e.g., heat-flux peaks, vibration bursts) coexist with slower trends (e.g., thermal equilibration, force ramps): early layers capture local transients, while deeper layers integrate over longer windows. Hence, shallow 1D-CNNs with small kernels are used as convolutional encoders in Models 1 and 3.

\subsection{Integrated Gradients (IG)}

To interpret each model, we employed Integrated Gradients (IG)—an explainable AI method that quantifies the contribution of each input feature to the model's output~\cite{Sundararajan2017IG}. For a scalar output $F(\mathbf{x})$ and baseline $\mathbf{x}'$ (a zero vector or matrix of the same shape as $\mathbf{x}$), the importance of feature $i$ is computed by integrating the gradient of $F$ along the linear path from baseline to input:

\begin{equation}
\mathrm{IG}_i(\mathbf{x}) = (x_i - x'_i) \int_{\alpha=0}^{1} \frac{\partial F(\mathbf{x}' + \alpha(\mathbf{x} - \mathbf{x}'))}{\partial x_i} \, d\alpha
\label{eq:ig}
\end{equation}

In practice, this integral is approximated via a Riemann sum over $N$ steps~\cite{Sundararajan2017IG}. For multivariate time series with channels $c$ and time steps $t$, IG produces a two-dimensional attribution map $\mathrm{IG}_{t,c}$. These attributions were further refined using Temporal Saliency Rescaling (TSR), which first identifies relevant time steps and then evaluates feature importance within them, reducing attribution dilution across multivariate time-series inputs~\cite{ismail2020_tsr}. The resulting attributions can then be summarized over specific time phases or channel groups.

\section {Models}

\subsection{Model~1: Signal to Adjective Rating Distributions}

This model estimates psychophysical adjective distributions from finger–surface interaction signals by mapping tactile signals acquired through three exploratory actions (T: thermal/static contact, S: sliding, and P: pressing) onto a discrete rating distribution spanning the 1–15 psychophysical scale. 

For each exploratory action $e \in \{\mathrm{T}, \mathrm{S}, \mathrm{P}\}$, we compute an embedding vector
\[
\mathbf{f}_{e} = E_{e}(\mathbf{x}^{(e)}) \in \mathbb{R}^{64},
\]
where $\mathbf{x}^{(e)} \in \mathbb{R}^{T_e \times C_e}$ is the input sequence for exploratory action $e$. The corresponding encoder $E_{e}$ consists of a stack of convolutional layers with activation functions, batch normalization, max-pooling, and dropout, followed by a flattening operation and a final dense layer. The depth of each encoder varies by action, with the sliding and pressing branches comprising more convolutional blocks than the thermal branch, reflecting differences in the temporal complexity of each signal type.

Then the action embeddings are concatenated into a fused representation
\begin{equation}
\mathbf{f} = [\mathbf{f}_{\mathrm{T}};\,\mathbf{f}_{\mathrm{S}};\,\mathbf{f}_{\mathrm{P}}],
\label{eq:model1_fused}
\end{equation}
which is passed through a shared fusion block. On top of this shared representation, Model~1 uses one output head per adjective pair. Each head consists of a Dense(64) layer with ReLU activation followed by dropout, and a final Dense(15) softmax layer that outputs the predicted rating distribution $\hat{\mathbf{q}}^{(a)} \in \Delta^{15}$.

This multi-head late-fusion design allows each action encoder specialize to its own signal characteristics (e.g., thermal transients, vibration/friction structure, indentation dynamics), while the shared fusion block learns how to combine these signals to predict perceived attributes.

Since Model~1 outputs a rating distribution over a 15-bin scale for each adjective pair, the training objective is cast as a distribution-matching problem. Accordingly, we adopt the Kullback–Leibler divergence (KLD) as both the loss function and the evaluation criterion.

\subsection{Model~2: Adjective-to-Material Classifier}
This model leverages the psychophysical adjective distributions predicted by Model~1 to classify materials. Specifically, for each sample, Model~1 produces a 15-bin distribution for every adjective pair; these are then concatenated into a single feature vector
\[
\mathbf{z} = [\hat{\mathbf{q}}^{(\text{hot--cold})};\ \hat{\mathbf{q}}^{(\text{hard--soft})};\ \hat{\mathbf{q}}^{(\text{rough--smooth})};\ \hat{\mathbf{q}}^{(\text{sticky--slippery})}] \in \mathbb{R}^{60}.
\]
This vector is fed into Model~2, a compact multilayer perceptron (MLP) consisting of a single dense hidden layer (default 32 units), batch normalization, dropout, and a softmax output layer for the seven material classes.

\subsection{Model 3: Signal-Based Material Classifier}
Model~3 uses action-specific encoders and a late-fusion block analogous to those in Model~1, but replaces the four adjective-specific distribution heads with a single material-classification head. The fused representation is passed through a dense layer, batch normalization, and dropout, followed by a softmax output layer over the seven material classes. The goal is to map tactile signals  acquired during static contact, sliding, or pressing directly to a
material class, without the intermediate adjective-attribute stage used in Models~1 and~2.

\section{Results and Discussion}

\subsection{Performance of Model~1} 

Table~\ref{tab:combined_results} (left) summarizes Model~1's performance in predicting probability distributions for each adjective pair, where lower KL divergence indicates better performance. Several observations emerge: the four adjective pairs are not equally predictable, with hot--cold being the most predictable pair and hard--soft the least predictable given the available data. These results suggest that the model's learned representations preferentially capture thermal perceptual structure over compliance-related structure. 

\begin{table}[b]
\centering
\caption{Model performance: Kullback–Leibler divergence (KLD) for Model 1 (left), F1 scores for Model 2 (middle), and F1 scores for Model 3 (right). HC: Hot--Cold, HS: Hard--Soft, RS: Rough--Smooth, SS: Sticky--Slippery.}
\label{tab:combined_results}
\begin{tabular}{lc|lc|lc}
\toprule
\multicolumn{2}{c|}{Model 1} & \multicolumn{2}{c|}{Model 2} & \multicolumn{2}{c}{Model 3} \\
\cline{1-2} \cline{3-4} \cline{5-6}
Adj. & \multicolumn{1}{c|}{KLD} & Class & \multicolumn{1}{c|}{F1 score} & Class & \multicolumn{1}{c}{F1 score} \\
\midrule
HC    & $0.29  \pm 0.04$ & Foam    & $0.54 \pm 0.34$ & Foam    & $0.84 \pm 0.10$ \\
HS    & $0.37  \pm 0.02$ & Metal   & $0.90 \pm 0.22$ & Metal   & $1.00 \pm 0.00$ \\
RS  & $0.35  \pm 0.02$ & Paper   & $0.36 \pm 0.21$ & Paper   & $0.80 \pm 0.09$ \\
SS & $0.33  \pm 0.02$ & Sandp.  & $0.16 \pm 0.36$ & Sandp.  & $0.87 \pm 0.18$ \\
All            & $0.34  \pm 0.02$ & Fabric  & $0.80 \pm 0.04$ & Vinyl   & $0.85 \pm 0.15$ \\
               &                   & Vinyl   & $0.07 \pm 0.15$ & Wood    & $0.84 \pm 0.12$ \\
               &                    & Wood    & $0.42 \pm 0.28$ & Fabric  & $1.00 \pm 0.00$ \\
               &                    & All     & $0.46 \pm 0.14$ & All     & $0.89 \pm 0.02$ \\
\bottomrule
\end{tabular}
\end{table}

\begin{figure}[t]
  \centering
  \begin{minipage}[t]{0.5\linewidth}
    \centering
    \includegraphics[width=\linewidth, trim={40 60 60 3.5cm}, clip]{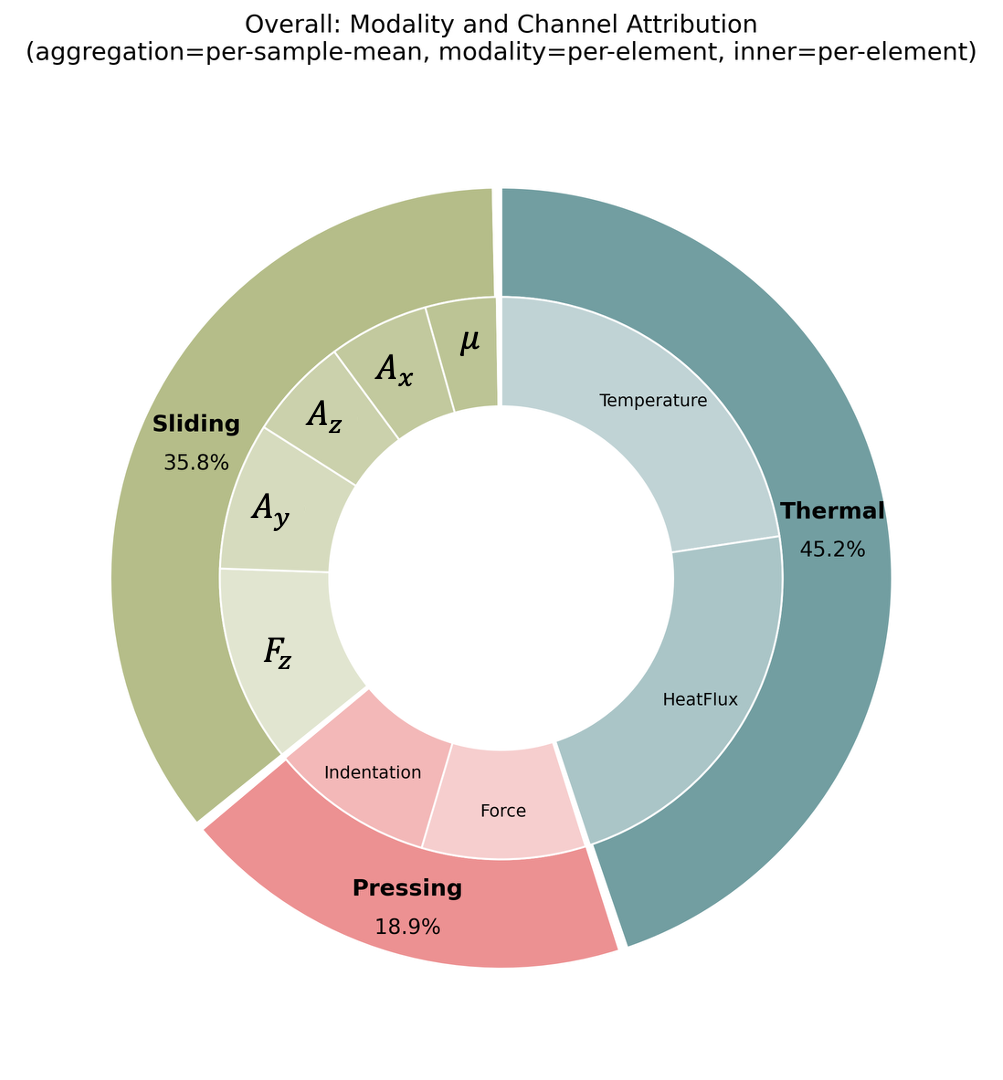}
    \caption*{\textbf{(a)} Hot--Cold} 
    \label{fig:m1_donut_hotcold_test}
  \end{minipage}\hfill
  \begin{minipage}[t]{0.5\linewidth}
    \centering
    \includegraphics[width=\linewidth, trim={60 60 60 3.5cm}, clip]{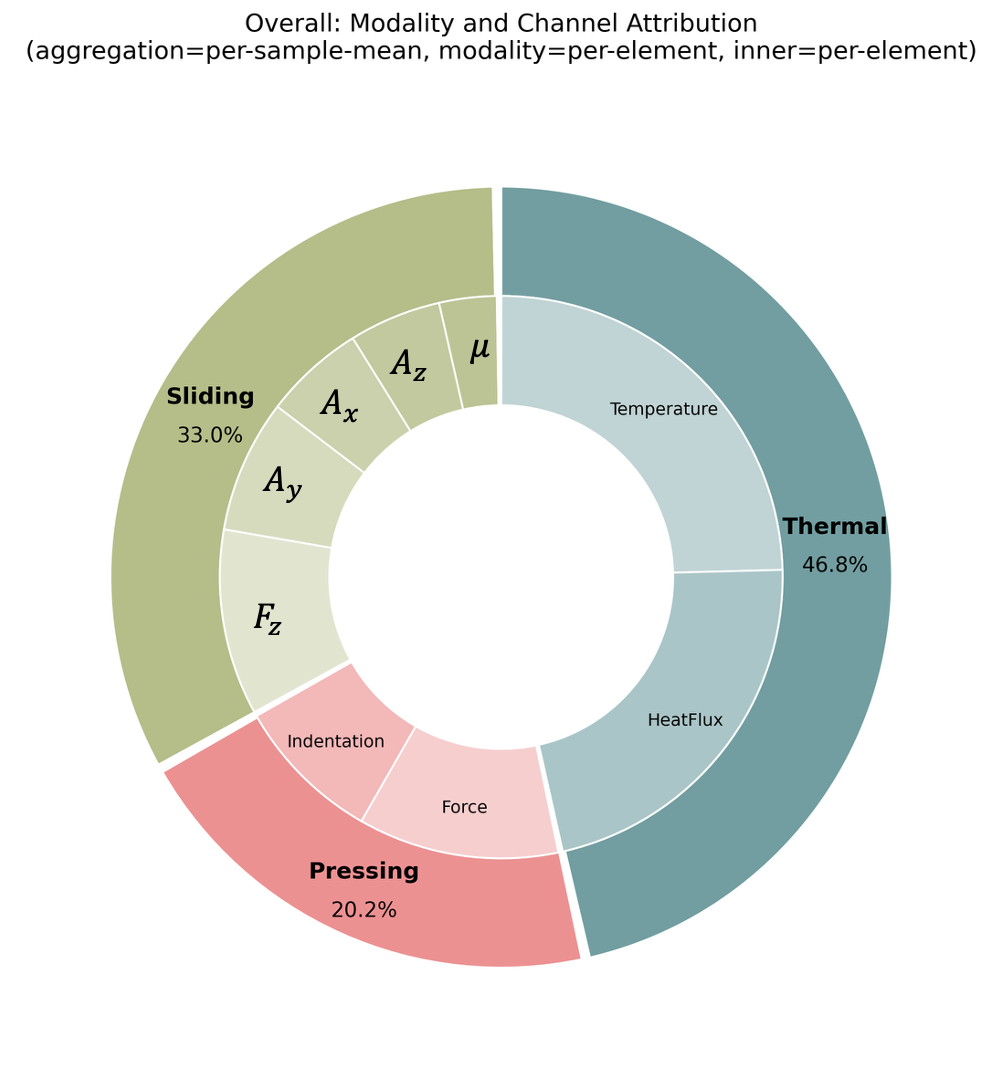}
    \caption*{\textbf{(b)} Hard--Soft}
    \label{fig:m1_donut_hardsoft_test}
  \end{minipage}
  \begin{minipage}[t]{0.5\linewidth}
    \centering
    \includegraphics[width=\linewidth, trim={60 60 60 3cm}, clip]{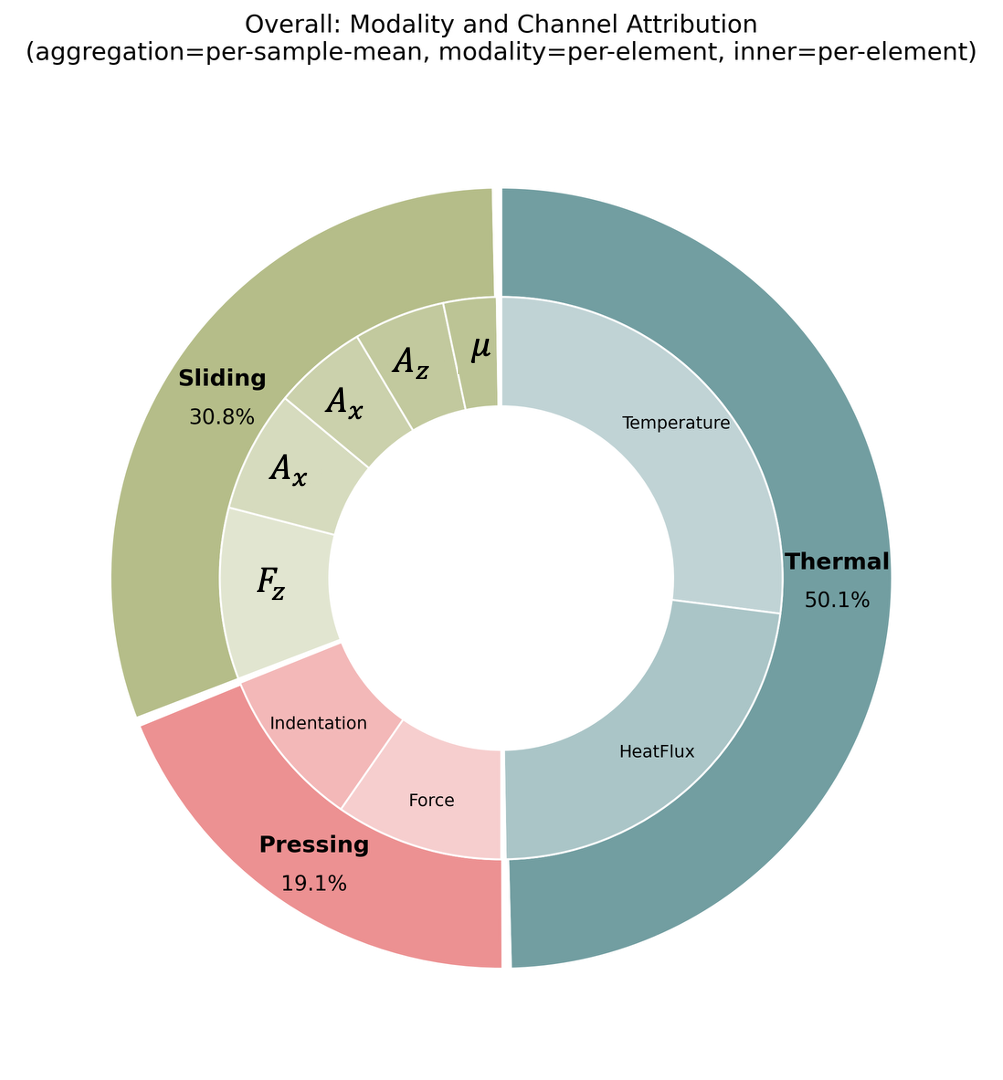}
    \caption*{\textbf{(c)} Rough--Smooth}
    \label{fig:m1_donut_roughsmooth_test}
  \end{minipage}\hfill
  \begin{minipage}[t]{0.5\linewidth}
    \centering
    \includegraphics[width=\linewidth, trim={60 60 60 3cm}, clip]{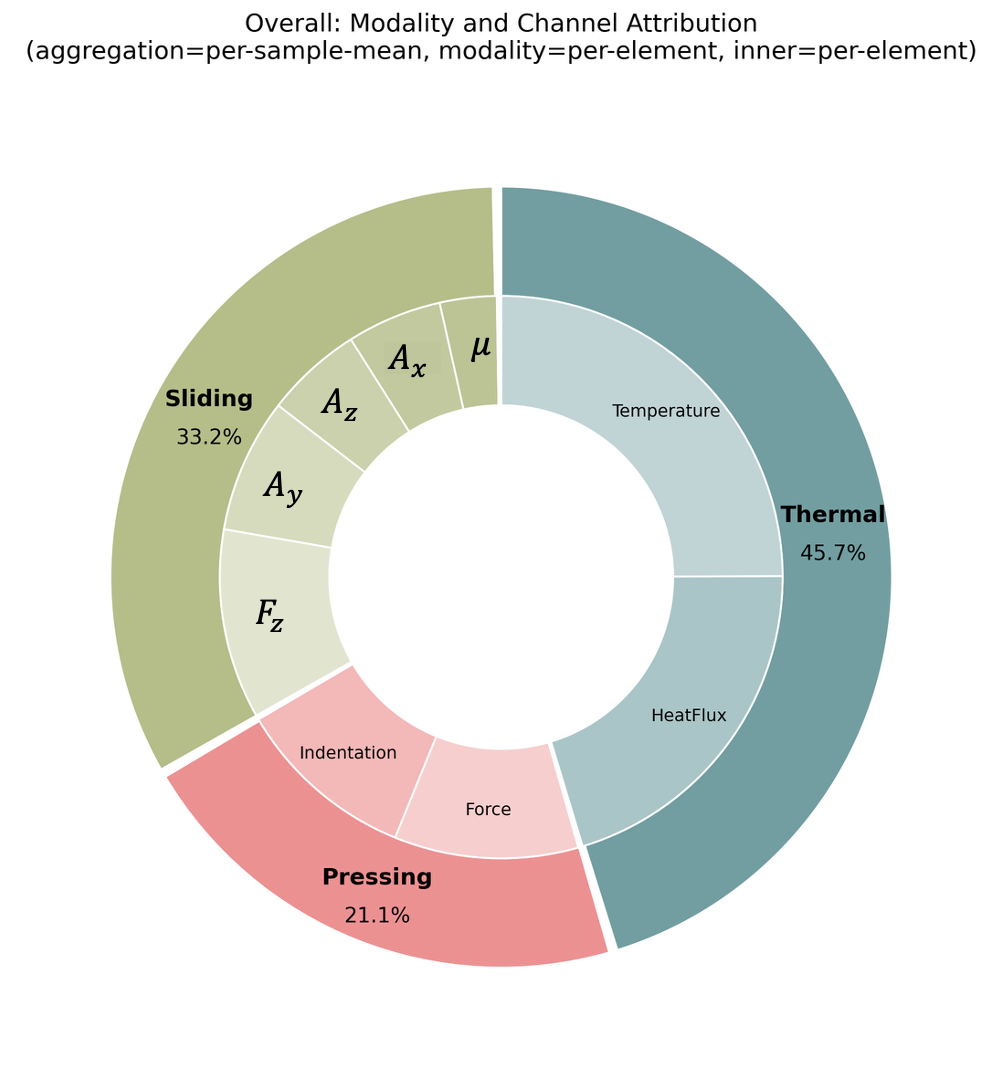}
    \caption*{\textbf{(d)} Sticky--Slippery}
    \label{fig:m1_donut_stickyslippery_test}
  \end{minipage}
  \caption{Model~1 interpretability results. Action-level attributions are shown in the outer ring and channel-level attributions in the inner ring for each adjective pair. As the static contact action includes only thermal measurements, it is labeled as "Thermal" for clarity. Here, $F_z$, $A_x$, $A_y$, $A_z$, and $\mu$ denote the applied normal force during sliding, recorded accelerations along the $x$, $y$, and $z$ axes, and the friction coefficient, respectively. }
  \label{fig:m1_donut_test_grid}
\end{figure}

The interpretability results for Model~1 (Fig.~\ref{fig:m1_donut_test_grid}) indicate a clear and consistent action hierarchy across adjective pairs. Thermal cues account for approximately half of the total attribution mass, dominating over the ones acquired from sliding (one-third) and pressing (one-fifth) actions. Subtle shifts emerge between specific pairs: rough--smooth draws disproportionately on thermal (static contact) data, while sticky--slippery exhibits the strongest dependence on pressing. Within each action, the inner ring identifies the most diagnostically informative interaction channels, with all channels contributing to some degree to the model's decisions. Interaction phases, which are omitted from the figures for clarity, follow a consistent pattern: for thermal attributions transient contact phases---specifically the peak and half-equilibration stages---are favored over steady contact; for pressing, the plateau and lift-off phases outweigh early loading; and for sliding, salience is greatest in the low-force regime and across low-to-mid speeds, consistent with the limited force–speed envelope of the dataset.

\subsection{Performance of Model 2} 

\begin{figure}[t]
  \centering
  \begin{minipage}[t]{1\linewidth}
    \centering
    \includegraphics[width=1\linewidth,trim={20 10 10 25},clip]{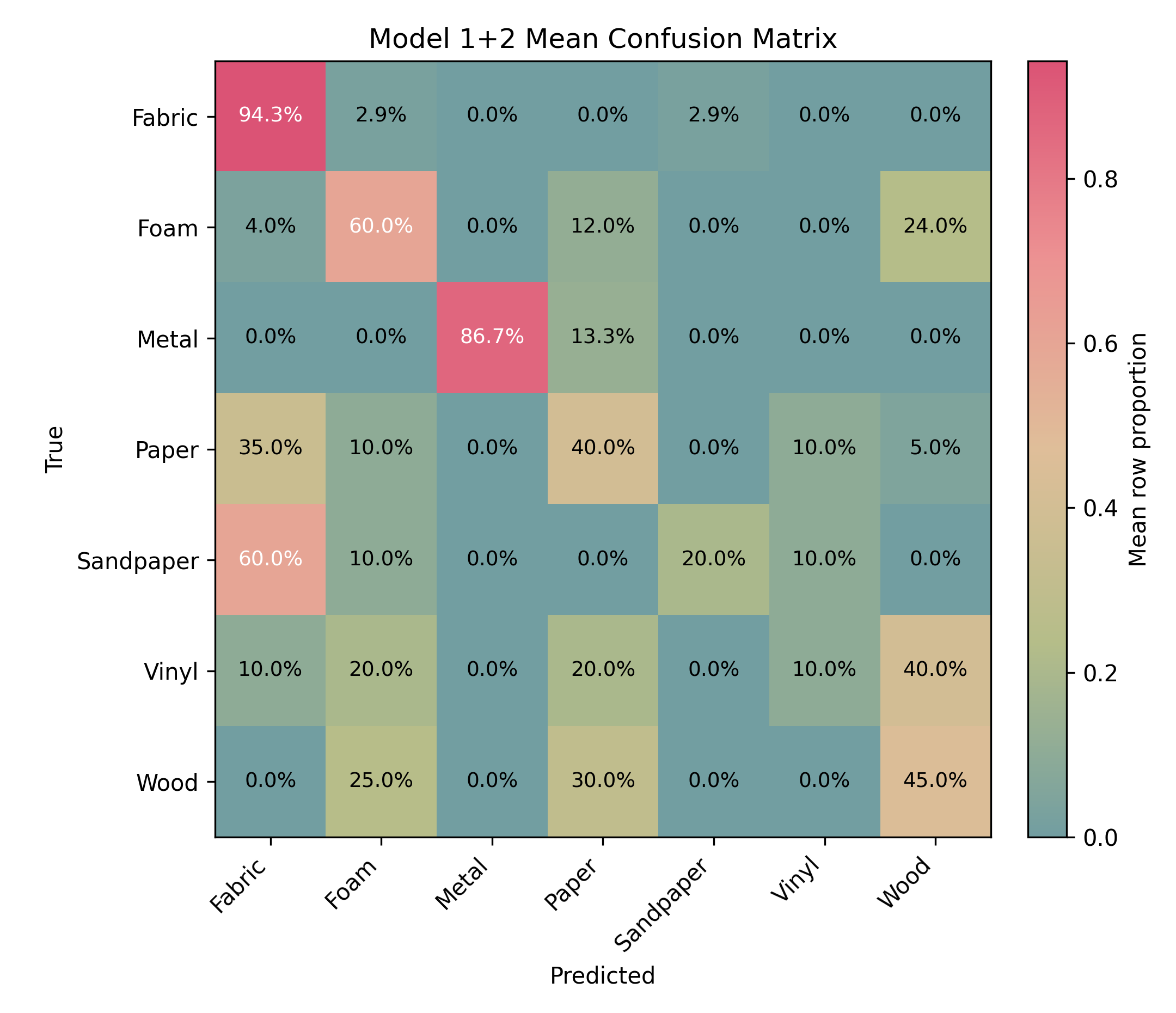}
    \caption*{(a) Model 2}
  \end{minipage}

  \vspace{0.5em}

  \begin{minipage}[t]{1\linewidth}
    \centering
    \includegraphics[width=1\linewidth,trim={20 10 10 23},clip]{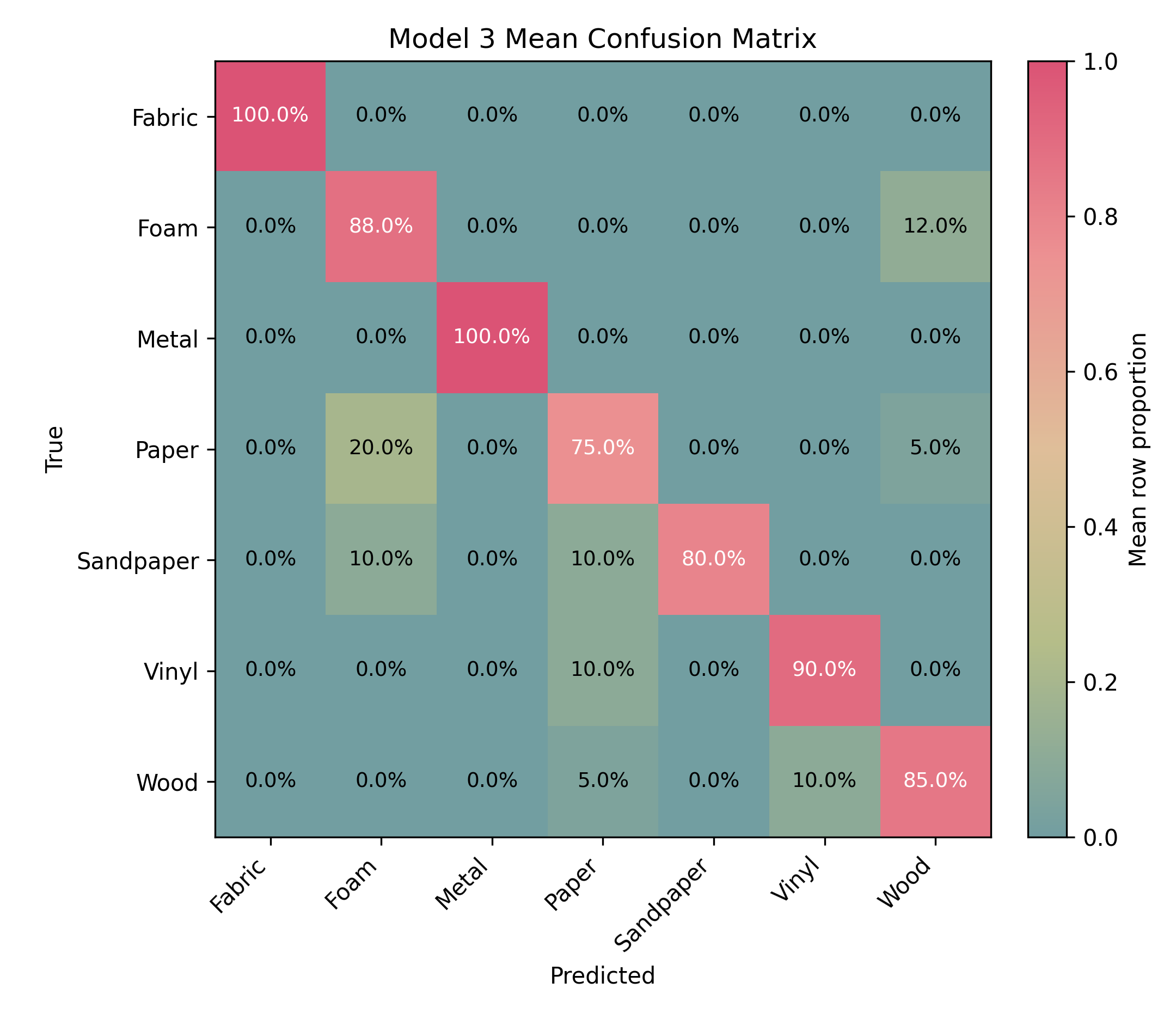}
    \caption*{(b) Model 3}
  \end{minipage}
  \caption{Mean confusion matrix for (a) Model~2 and (b) Model~3, row-normalized and aggregated across folds.}
  \label{mean_confusion_matrix}
\end{figure}

Table~\ref{tab:combined_results} (middle) summarizes F1 scores for Model~2 on the test set, for each material class as well as their overall performance. A higher F1 score reflects a better trade-off between precision and recall, indicating better performance. The moderate overall score and the spread in per-class F1 scores together indicate limited and uneven classification capability: Fabric and Metal are recognized reasonably well, while Paper, Sandpaper, and Vinyl remain challenging. The confusion matrix in Fig.~\ref{mean_confusion_matrix} (a) diagnoses the source of these weaknesses: Paper and Sandpaper are commonly confused with Fabric, and Vinyl is predominantly mislabeled as Wood.

\begin{figure}[htbp]
  \centering
  \begin{minipage}[t]{1\linewidth}
    \centering
    \includegraphics[width=0.8\linewidth, trim={0 0 430 1.0cm}, clip]{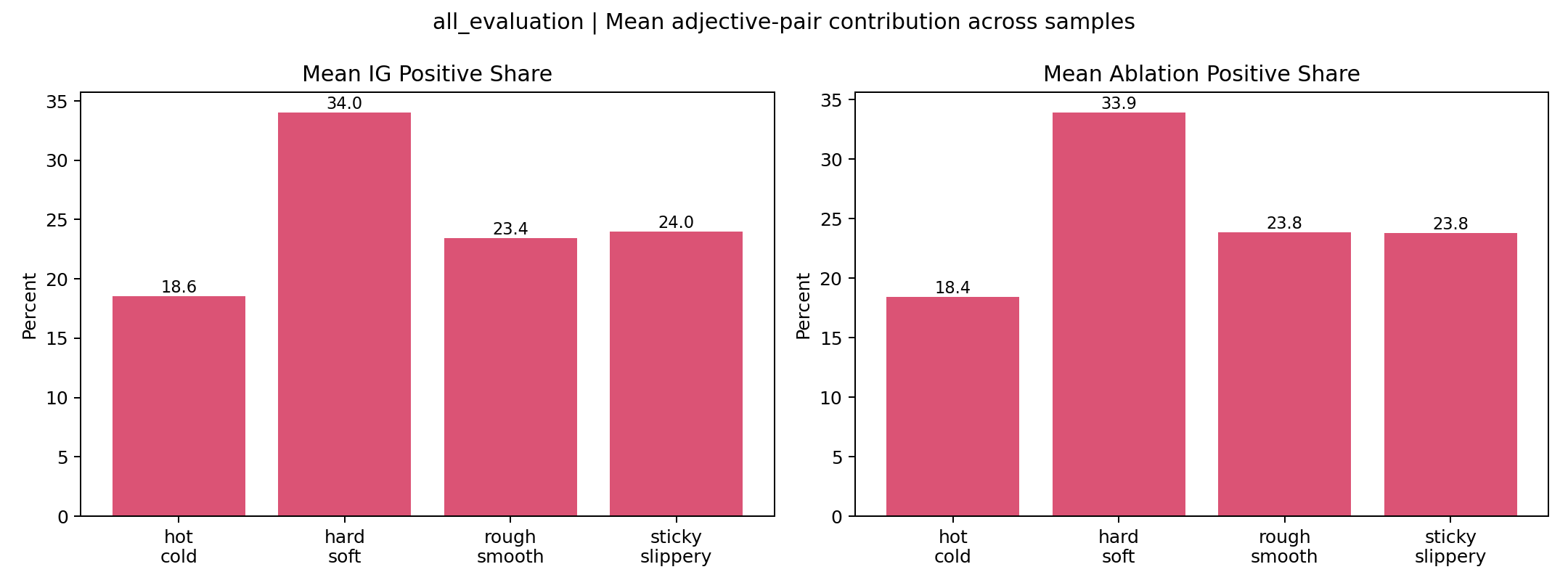}
    \caption*{(a) Overall attribution share per adjective pair}
  \end{minipage}

  \vspace{0.5em}

  \begin{minipage}[t]{1\linewidth}
    \centering
    \includegraphics[width=0.95\linewidth, trim={0 0 0 0}, clip]{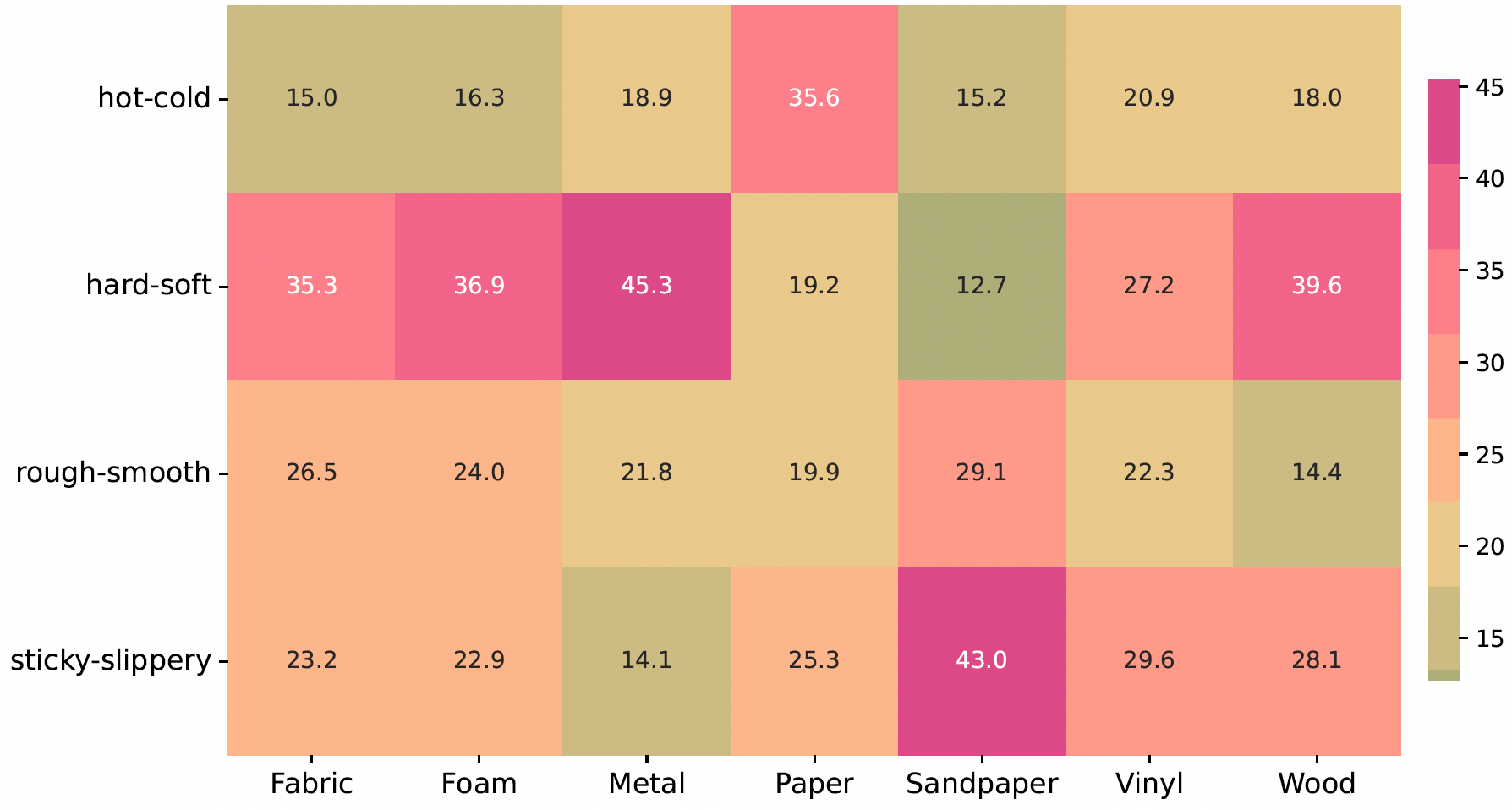}
    \caption*{(b) Attribution share per adjective pair by material class}
  \end{minipage}

  \caption{Model~2 interpretability results.}
  \label{fig:model2_interpretability_combined}
\end{figure}

Interpretability results for Model~2 show that the hard--soft adjective pair exerted the greatest influence on material classification, accounting for roughly 34.0\% of the positive IG share across all adjective pairs. The remaining attribution was distributed more evenly among rough--smooth (23.4\%), sticky--slippery (24.0\%), and hot--cold (18.6\%), as illustrated in Fig.~\ref{fig:model2_interpretability_combined} (a). At the individual material level (Fig.~\ref{fig:model2_interpretability_combined} (b)), this hierarchy remains broadly consistent: hard--soft dominates for most material classes, contributing especially strongly to the prediction of Metal, Wood, Foam, and Fabric. Sticky--slippery and hot--cold, in turn, contribute most prominently to predicting Sandpaper and Paper, respectively.

\begin{figure}[htbp]
  \centering
  \begin{minipage}[t]{0.5\linewidth}
    \centering
    \includegraphics[width=\linewidth, trim={60 60 60 5cm}, clip]{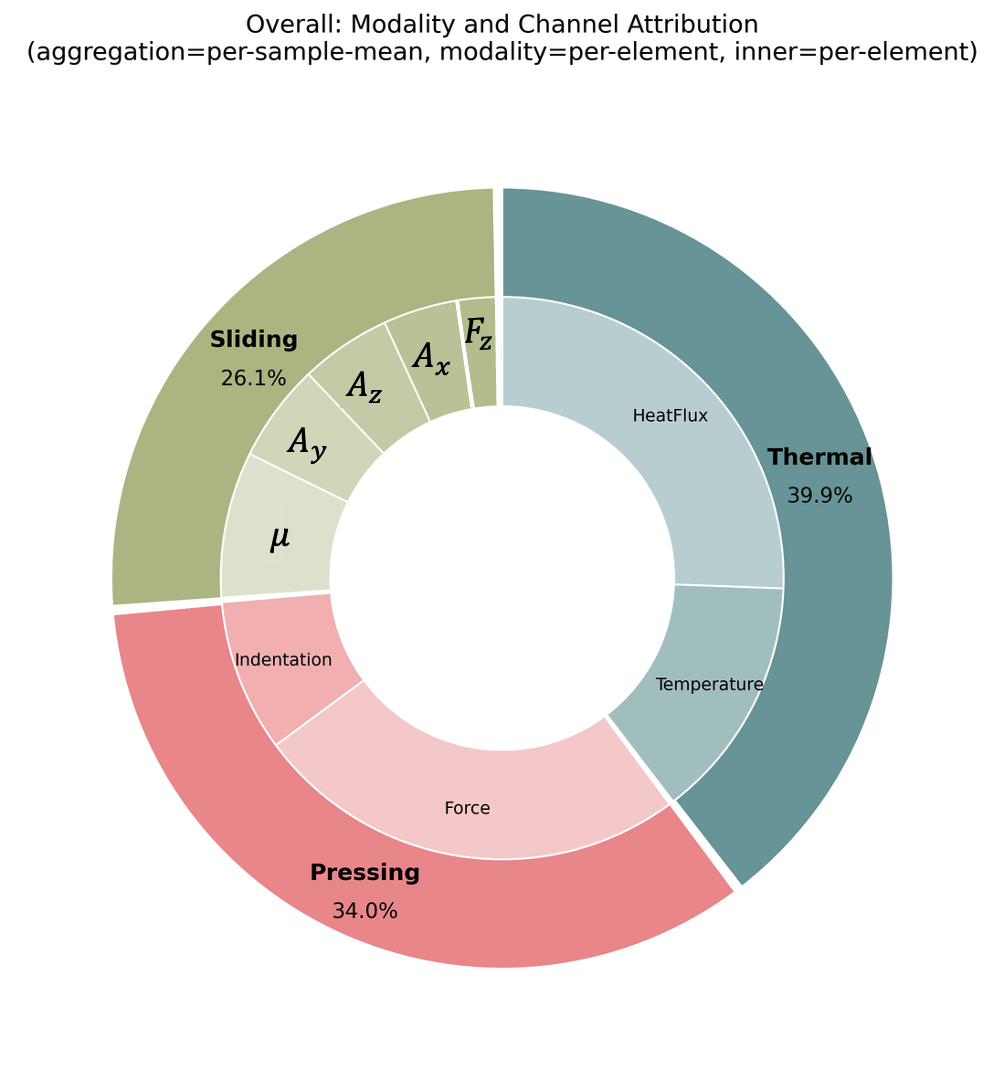}
    \caption*{(a) Overall}
    \label{fig:m3_donut_overall_channels_test}
  \end{minipage}\hfill
  \begin{minipage}[t]{0.5\linewidth}
    \centering
    \includegraphics[width=\linewidth, trim={60 60 60 5cm}, clip]{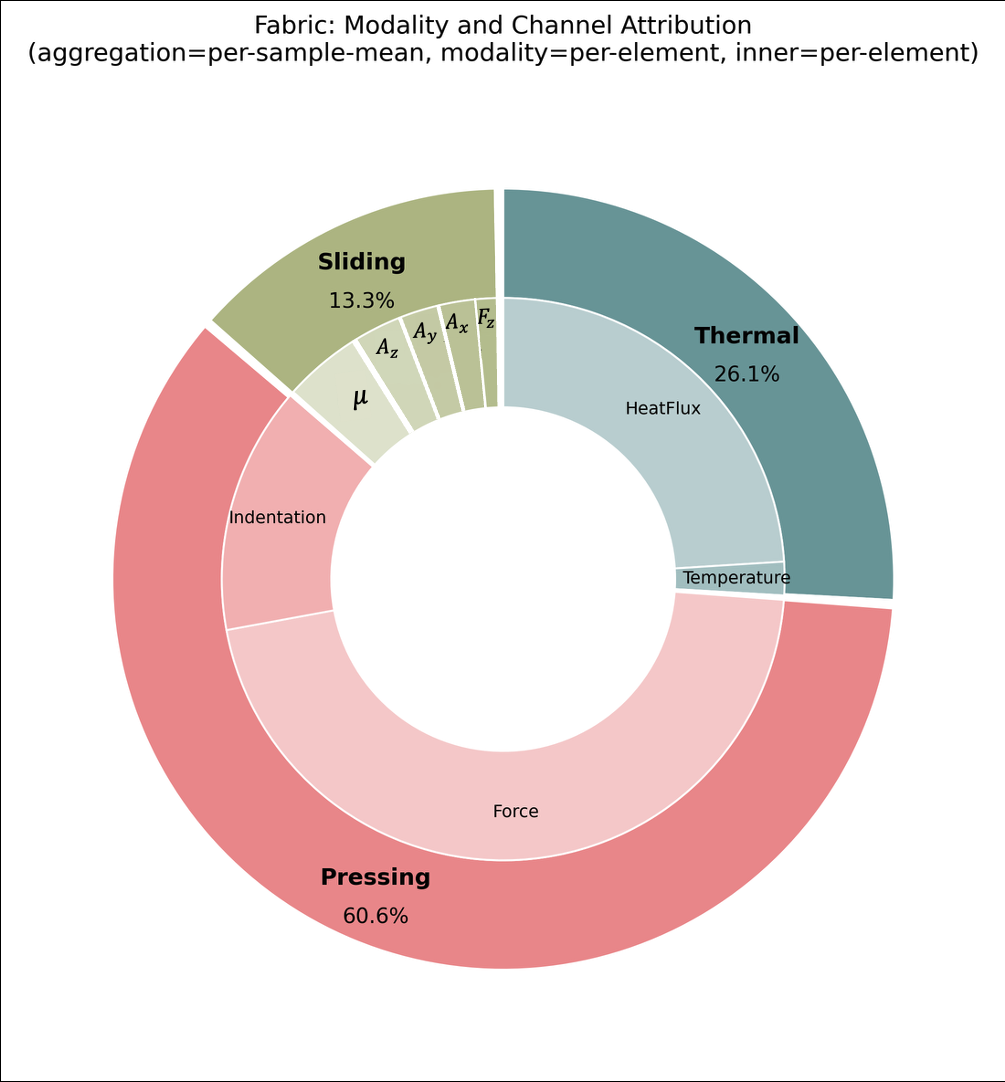}
    \caption*{(b) Fabric}
    \label{fig:m3_donut_fabric_channels_test}
  \end{minipage}

  \vspace{0.5em}

  \begin{minipage}[t]{0.5\linewidth}
    \centering
    \includegraphics[width=\linewidth, trim={60 60 60 5cm}, clip]{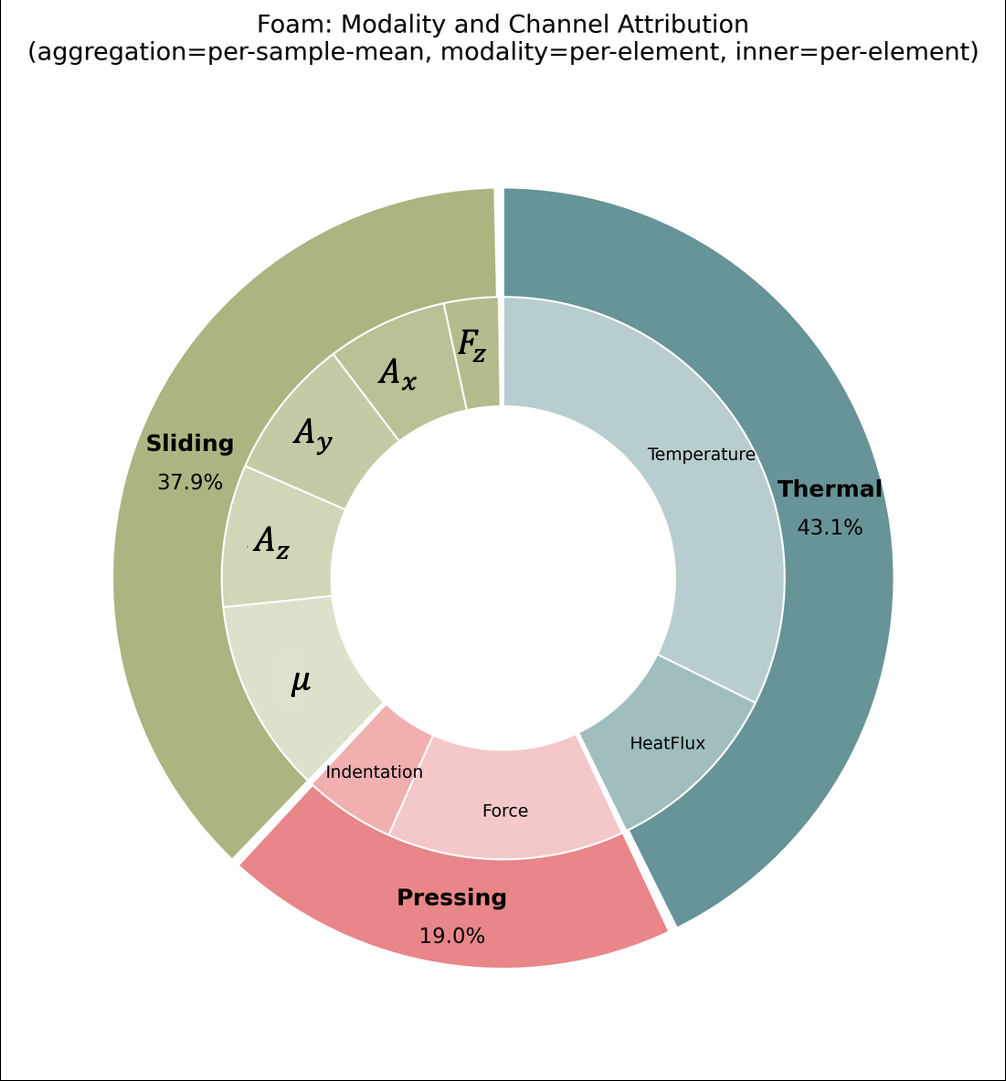}
    \caption*{(c) Foam}
    \label{fig:m3_donut_foam_channels_test}
  \end{minipage}\hfill
  \begin{minipage}[t]{0.5\linewidth}
    \centering
    \includegraphics[width=\linewidth, trim={60 60 60 5cm}, clip]{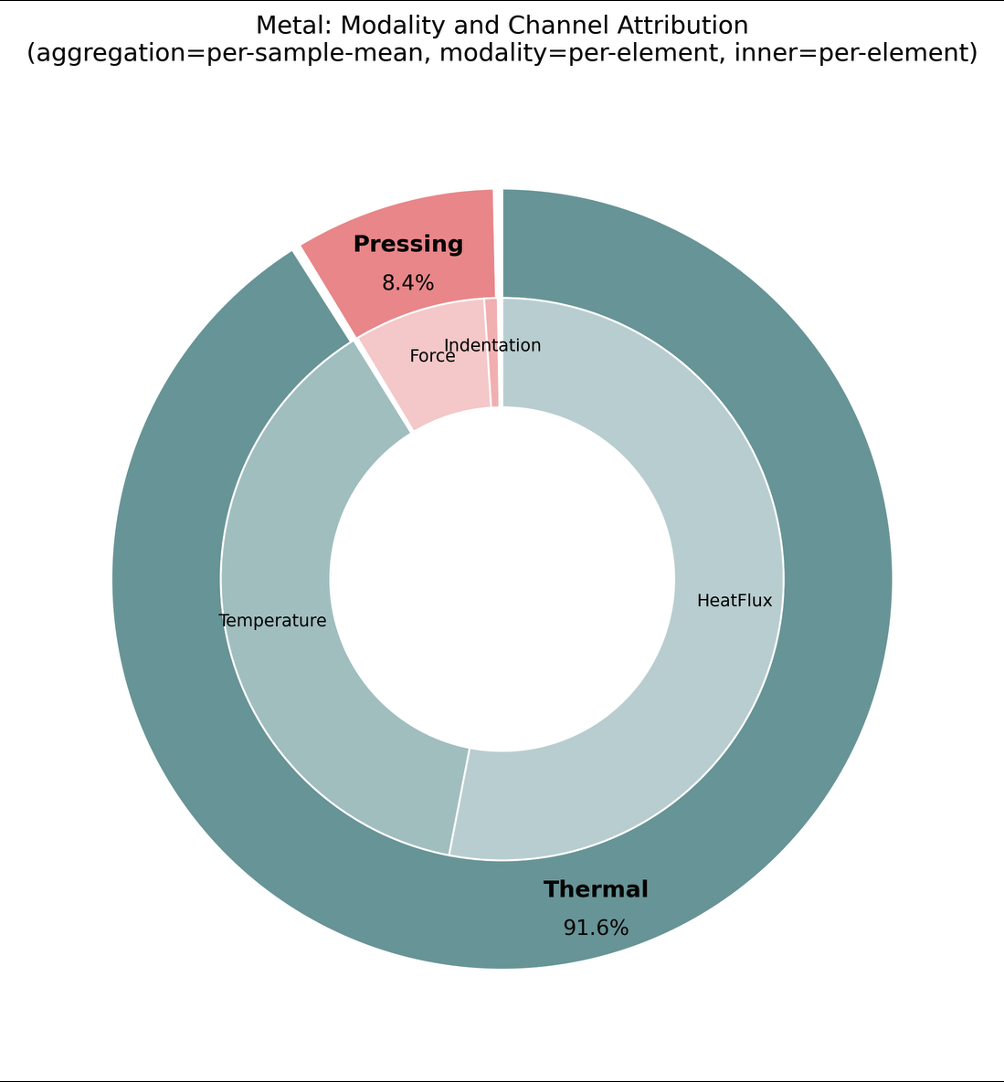}
    \caption*{(d) Metal}
    \label{fig:m3_donut_metal_channels_test}
  \end{minipage}

  \vspace{0.5em}

  \begin{minipage}[t]{0.5\linewidth}
    \centering
    \includegraphics[width=\linewidth, trim={60 60 60 5cm}, clip]{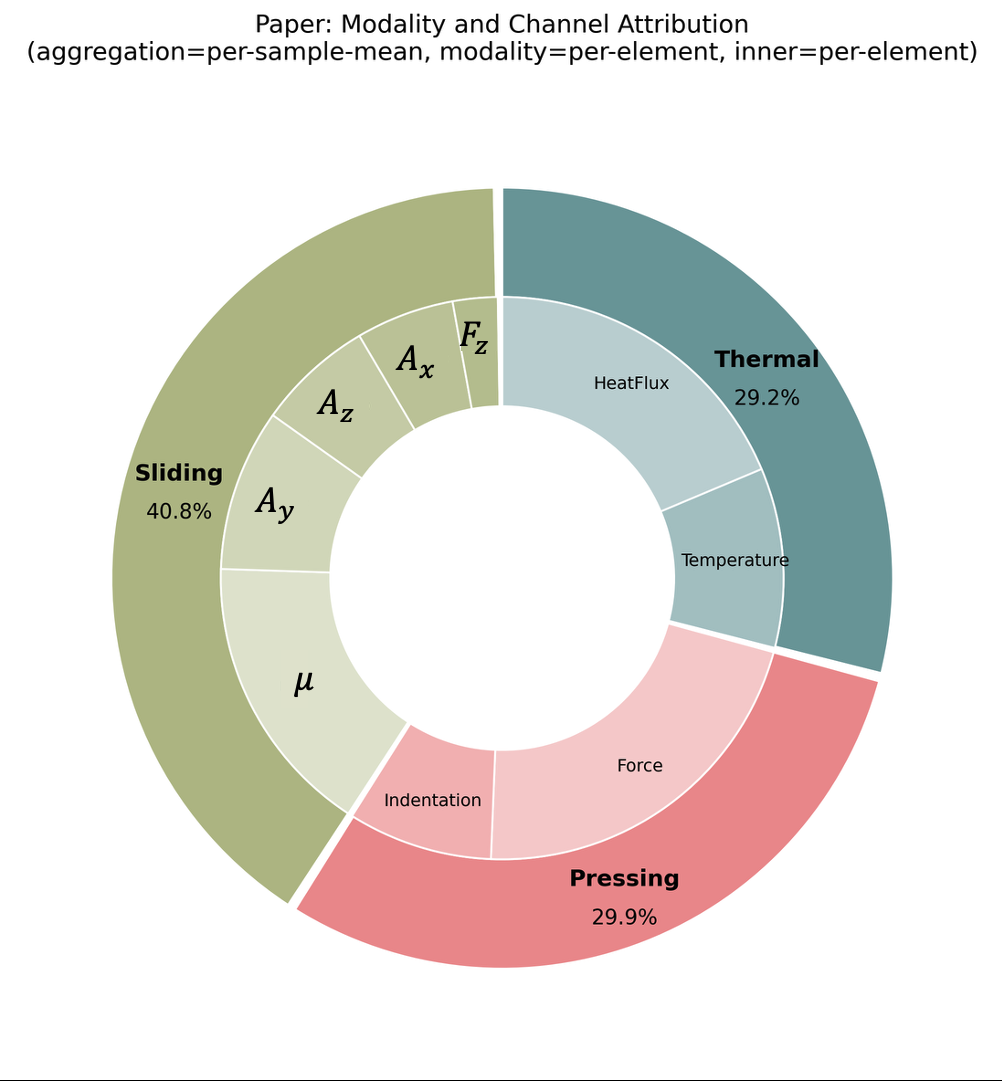}
    \caption*{(e) Paper}
    \label{fig:m3_donut_paper_channels_test}
  \end{minipage}\hfill
  \begin{minipage}[t]{0.5\linewidth}
    \centering
    \includegraphics[width=\linewidth, trim={60 60 60 5cm}, clip]{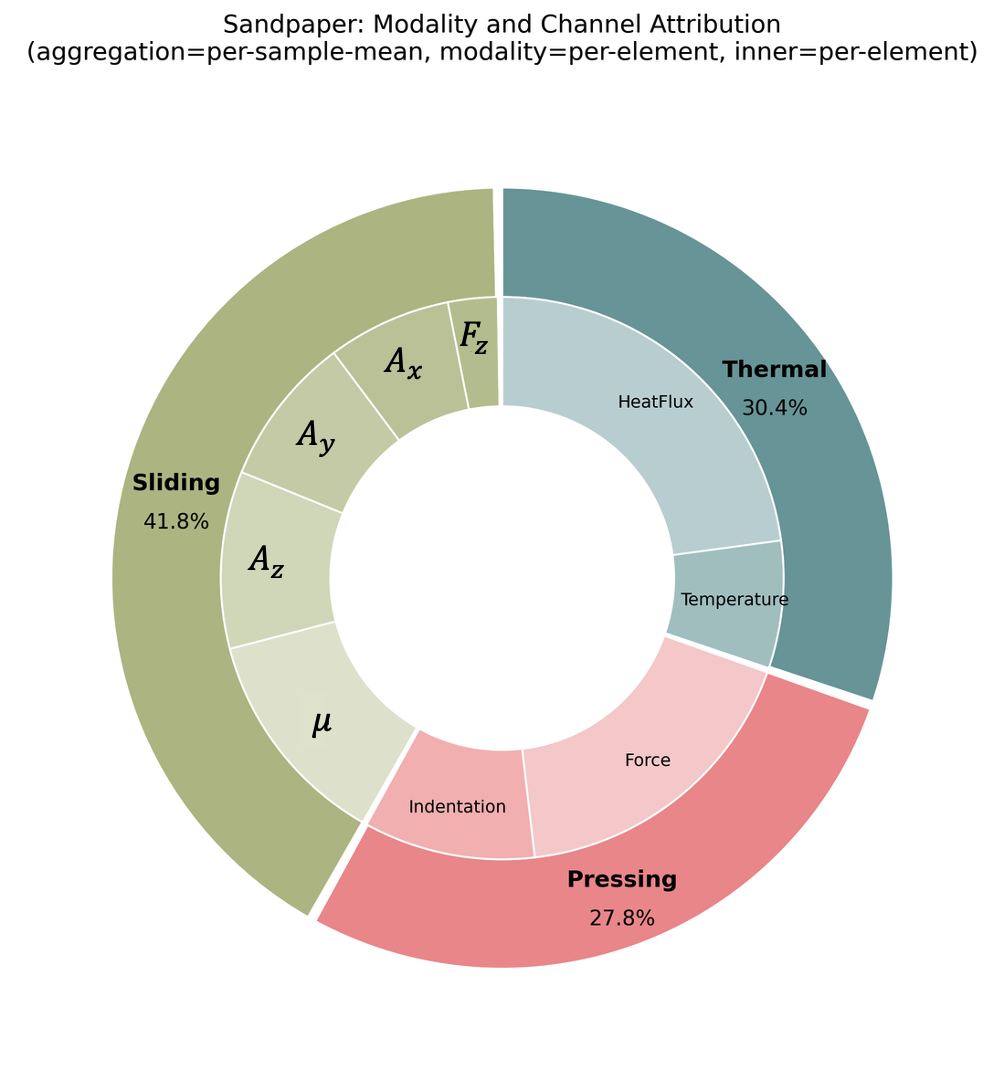}
    \caption*{(f) Sandpaper}
    \label{fig:m3_donut_sandpaper_channels_test}
  \end{minipage}

  \vspace{0.5em}

  \begin{minipage}[t]{0.5\linewidth}
    \centering
    \includegraphics[width=\linewidth, trim={60 60 60 5cm}, clip]{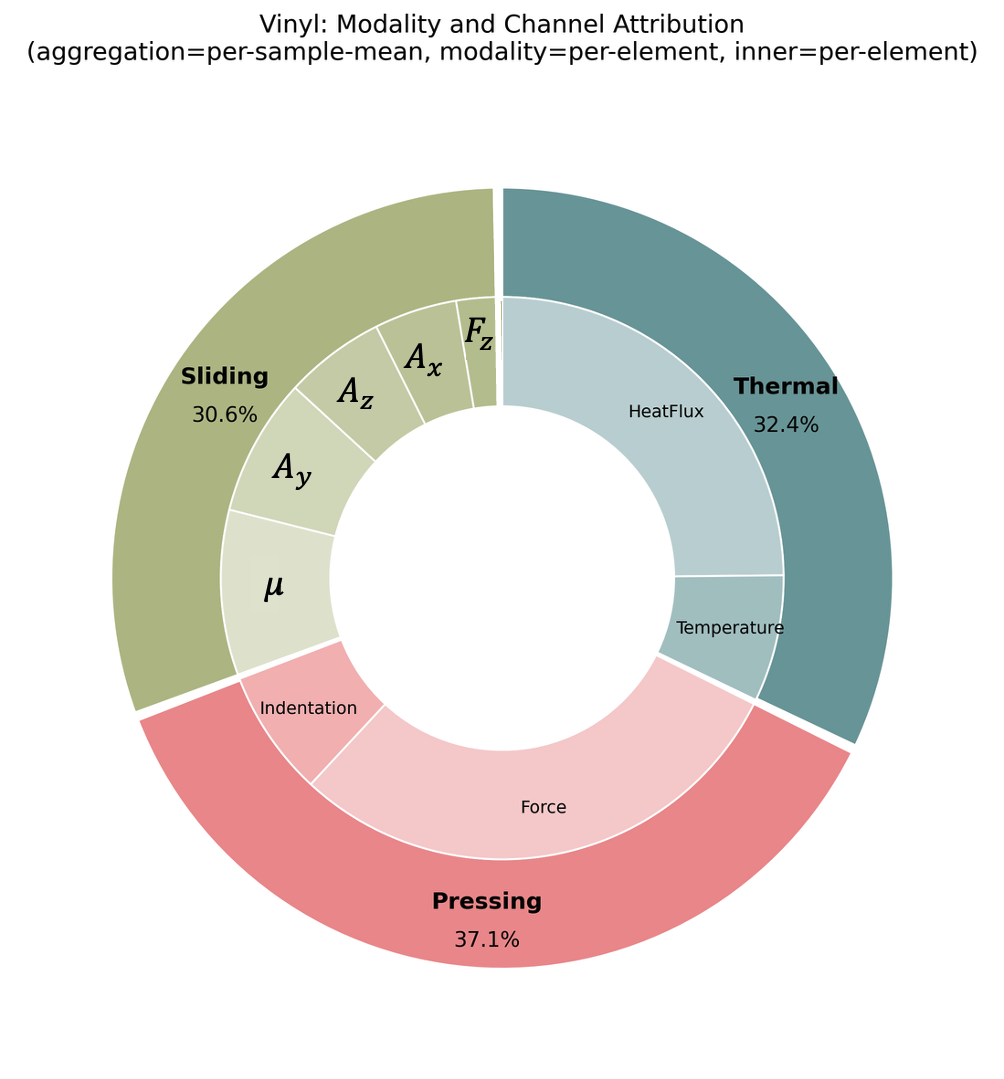}
    \caption*{(g) Vinyl}
    \label{fig:m3_donut_vinyl_channels_test}
  \end{minipage}\hfill
  \begin{minipage}[t]{0.5\linewidth}
    \centering
    \includegraphics[width=\linewidth, trim={60 60 60 5cm}, clip]{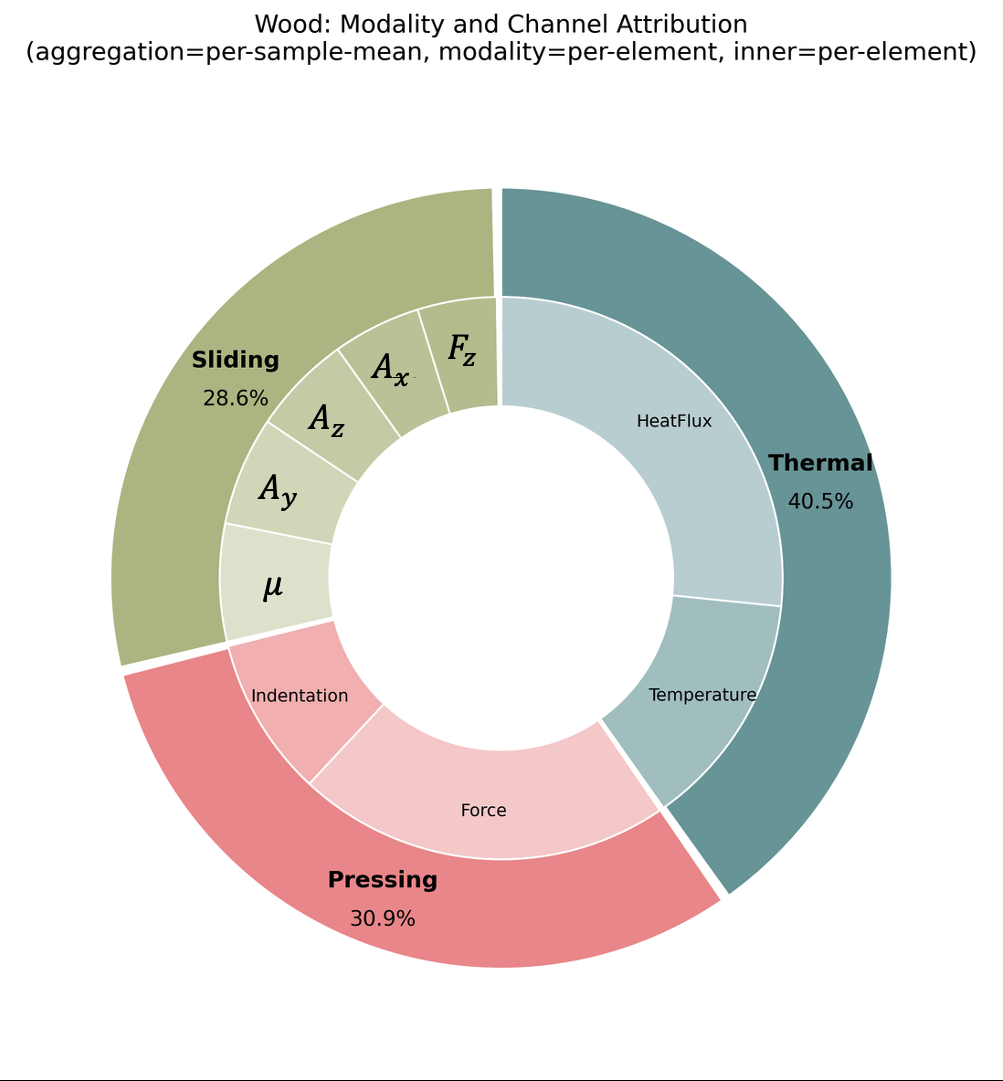}
    \caption*{(h) Wood}
    \label{fig:m3_donut_wood_channels_test}
  \end{minipage}

  \caption{Model~3 interpretability results. The outer ring shows the relative attribution share of static contact (thermal), pressing, and sliding actions; the inner ring shows the channel-level breakdown within each modality. Here, $F_z$, $A_x$, $A_y$, $A_z$, and $\mu$ denote the applied normal force during sliding, recorded accelerations along the $x$, $y$, and $z$ axes, and the friction coefficient, respectively. }
  \label{fig:m3_donut_all_classes_channels_test}
\end{figure}

\subsection{Performance of Model~3}

Model~3 performs material classification directly from tactile signals, without any intermediate step. The model achieves strong overall F1 performance (Table~\ref{tab:combined_results} (right)). Fabric and Metal are identified nearly perfectly, and Foam, Sandpaper, Vinyl, and Wood also yield high F1 scores. Paper is the most challenging class, carrying a 20\% probability of being confused with Foam (see the confusion matrix in Fig.~\ref{mean_confusion_matrix}(b)).

Model~3 exhibits a balanced interpretability profile across modalities (Fig.~\ref{fig:m3_donut_all_classes_channels_test}a). On the test set, the contribution shares are 39.9\% for thermal, 34.0\% for pressing, and 26.1\% for sliding. A closer examination of within-action patterns reveals how Model~3 utilizes the input signals. At the channel level, thermal attribution is dominated by heat flux rather than temperature; pressing is driven primarily by normal force rather than indentation; and sliding depends most strongly on the friction coefficient, $\mu$, and the band-pass IMU axes, with normal force $F_z$ contributing the least. At the phase level, thermal attribution concentrates in transient heat-transfer phases---particularly the peak and half-equilibration stages---rather than in steady-state contact. Pressing attribution is distributed across loading, plateau, and lift-off. Sliding attribution is dominated by low-force bins across multiple speed ranges.

A per-class donut chart in Fig.~\ref{fig:m3_donut_all_classes_channels_test} (b-h) reveals clearer specialization patterns. Metal is strongly thermal-driven, consistent with its very high classification performance. Foam relies heavily on sliding and thermal contributions, while Fabric is more pressing-driven. Paper, Wood, Vinyl, and Sandpaper exhibit more mixed action usage. This mixed profile likely contributes to confusion among these classes: as they cannot be separated by a single dominant action, the model must rely on a combination of signals that also appear in other materials.

\subsection{A Comparison Between Path~1 and Path~2}

Path~1 and Path~2 both perform material classification from interaction signals, but Path~1 includes an intermediate stage that Path~2 lacks. Based on the results in Table~\ref{tab:combined_results} (middle) and Table~\ref{tab:combined_results} (right), Path~2 outperforms Path~1. A likely explanation is that additional intermediate stages accumulate prediction errors, degrading final performance. In this case, the extra intermediate stage in Path~1 may account for its weaker classification results. This result suggests that achieving human-like behavior or decision-making in AI-based material classification is challenging, as additional intermediate stages must be accounted for.

The seven material classes can be divided into three categories based on their classification behavior. The first category consists of Fabric and Metal, which are easily classified regardless of the path taken. The second category contains only Vinyl, which is correctly classified only via Path~2. The remaining materials form the third category, as they can be reasonably classified by both paths.

Among the sensory modalities examined, thermal cues emerged as particularly informative for both adjective rating distributions in Path~1 and material classification in Path~2. This finding suggests that thermal information provides a salient and robust signal for differentiating materials within computational models.

\section{Conclusion}

In this work, we adopt a computational framework for modeling human material perception and recognition from multisensory touch data using deep learning, which eliminates the need for hand-crafted feature engineering. We develop three interconnected models that progressively map tactile information across representational levels: from low-level interaction signals, to perceptual attribute distributions, and finally to material categories. Model~1 captures the relationship between tactile signals and the psychophysical sensory attributes reported by human participants. Model~2 uses these predicted attribute distributions to classify materials in a manner that approximates human recognition. Model~3 bypasses the intermediate perceptual stage and classifies material types directly from tactile signals using deep learning. Together, these models enable a systematic examination of how different representations contribute to material recognition without making claims about the underlying biological or neural mechanisms of human perception.

The combination of deep learning and Integrated Gradients provides both strong classification accuracy and interpretability, revealing which sensory modalities---and which aspects of them---contribute most to the model's decisions. As such, it may inform the design of future tactile sensing systems and haptic interfaces that are more closely aligned with human perceptual behavior. These findings also indicate that deep learning models can approach near-perfect material classification when unconstrained by intermediate rules. Nevertheless, achieving human-like behavior or decision-making in AI-based material classification continues to pose a significant challenge, as it demands accounting for the intermediate stages inherent to human perception.

Among the sensory modalities examined, thermal cues emerged as particularly informative for both perceptual modeling and material classification. This finding aligns with prior work on human haptic perception and suggests that thermal information provides a salient and robust signal for differentiating materials within computational models~\cite{Baum2013_visual, okamoto2013_dimensions, zou2026, sanz2026_multimodal}. Despite their importance in perceptual modeling, thermal cues remain underrepresented in current robotic fingers and haptic display technologies, which predominantly focus on single-modality mechanical sensing ~\cite{lepora} and actuation~\cite{chen2026_review, biswas2019_emerging}. Our results suggest that incorporating sensors capable of encoding thermal information may improve robotic systems’ ability to approximate human-like material perception. Similarly, haptic interfaces could benefit from richer multisensory display capabilities, including thermal feedback, to better approximate real-world material interactions in virtual and digital environments~\cite{vardar2025_multimodal, Kodak2023, sanz2026_multimodal}.

\bibliography{citations}
\end{document}